\documentclass{article}
\usepackage{spconf,amsmath,graphicx,hyperref}

\usepackage{amsmath,amssymb}
\usepackage{graphicx}
\usepackage{booktabs}
\usepackage{algorithm}
\usepackage{algpseudocode}
\usepackage{tikz}
\usetikzlibrary{arrows.meta,positioning}
\usepackage{xcolor}
\usepackage{cite}
\usepackage{hyperref}
\hypersetup{colorlinks=true,linkcolor=blue,citecolor=blue,urlcolor=blue}
\usepackage{cite}
\usepackage{amsmath,amssymb,amsfonts}

\usepackage{graphicx}
\usepackage{textcomp}
\usepackage{xcolor}
\def\BibTeX{{\rm B\kern-.05em{\sc i\kern-.025em b}\kern-.08em
    T\kern-.1667em\lower.7ex\hbox{E}\kern-.125emX}}

\usepackage{appendix}
\usepackage{hyperref}
\usepackage{url}
\usepackage{float}
\usepackage[noabbrev,capitalize]{cleveref}

\usepackage{comment}
\usepackage{subcaption}
\usepackage{multirow}
\usepackage{multicol}
\usepackage{times}
\usepackage{array}
\usepackage{latexsym}
\usepackage{tabularx}
\usepackage{inconsolata}
\usepackage{enumitem}
\usepackage{listings}
\usepackage{placeins}
\usepackage{cuted}        
\usepackage{capt-of}      
\usepackage{booktabs}
\usepackage{makecell} 
\usepackage{dblfloatfix}
\usepackage{cuted}
\allowdisplaybreaks
\newcommand{\discogp}{\textsc{DiscoGP}}
\newcommand{\circuittree}{\textsc{CircuitTree}}

\title{CircuitTree: Composing Circuits into  A Syllable Transcriber}
\name{Jiankun Wei$^1$, Ewan Dunbar$^2$, Gerald Penn$^3$\thanks{This work was supported by [embargoed for anonymous review].}}
\address{$^{123}$Univerity of Toronto}
\begin{document}
%
\maketitle
\begin{abstract}
The goal of circuit discovery is to decompose a monolithic neural network into tiny modular subnetworks which are interpretable in the sense that each does an identifiable subtask. We develop a practical method for composing arbitrarily many circuits into a single model.  We propose a greedy, density-scaled composition algorithm that overcomes limitations of previous attempts to compose circuits by naive union, then generalize it to $N$ circuits via CircuitTree, a new data structure built on a $\Theta(N\log N)$ tree-compose schedule. We compose circuits pruned from HuBERT for classifying individual phonemes into a syllable classifier, a capability none of the 39 phoneme circuits has individually. Chaining CircuitTrees into a hierarchy-of-circuits pipeline, we transcribe open syllables at 64.8\% accuracy, within 1.2 points of a 66.0\% ceiling that shares no circuitry at all, with 99.5\% of circuit components shared. 
\end{abstract}
\section{Introduction}
\label{sec:intro}

Many researchers has proposed to view modern multi-task neural network models as the composition of large numbers of sub-networks (circuits) with independent capabilities. ~\cite{mondorf2025compositions, yu2025sheaf, merullo2024circuit, chen2026circuits, chowdhary2025kmshc, wei2026hubertcircuits}. Early work identified circuits manually \cite{wang2022interpretability}; automated methods soon followed ~\cite{conmy2023acdc}, reducing circuit discovery to a process of choosing a metric and a dataset to define a behaviour, and patching activations until the responsible components emerge.
\begin{figure}[t!]
    \centering
    \includegraphics[height=0.5\linewidth]{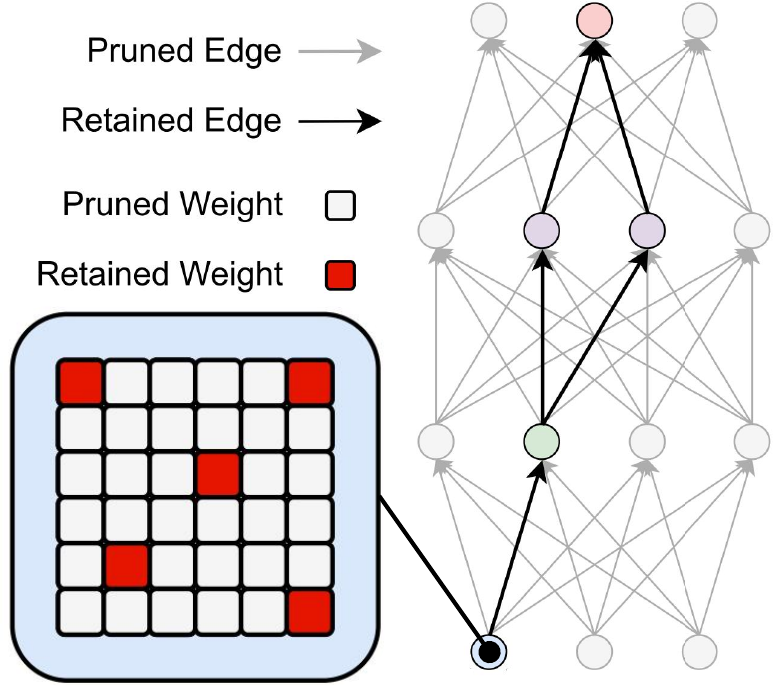}
    \caption{Circuit discovery of weight and edge circuits.}
    \label{fig:overview}
\end{figure}

While much research has proposed circuit discovery as a tool for deepening our understanding of black-box models, the ability to decompose models into small, modular sub-networks has practical implications.  Researchers  have envisioned circuit discovery as a way to ``re-engineer'' a network by decomposing the monolithic model into modular, Lego-like pieces, each responsible for a single task ~\cite{mondorf2025compositions, yu2025sheaf, pan2020decomposing, kingetsu2021modular, wei2026hubertcircuits}. The practical payoff of this decomposition is modularity. Once primitive circuits are identified, they can be joined to solve complex tasks without having to rediscover every possible combination from scratch. Merullo, Eickhoff, and
Pavlick~\cite{merullo2024circuit} found that a circuit discovered for one task in GPT-2 is substantially reused for a related one, showing that a small hand-designed patch to four attention heads is sufficient to change the model's behaviour. Mondorf et al. ~\cite{mondorf2025compositions} went further, posing general circuit composability as a fundamental open question and combining circuits for ten string-edit operations through mask-level set operations. However, their experiments were limited to synthetic toy models generated via Tracr ~\cite{lindner2023tracr}. Most recently, Yu et al.~\cite{yu2025sheaf} discovered that composing two independently discovered circuits pruned from GPT-2 for different tasks by simple mask union largely preserves functional performance.

This is a compelling vision, but the evidence for it remains preliminary. Existing composition results stop at a pair of circuits, or at best a handful, and success is measured by whether the composed circuit still performs the individual tasks it was pruned on. Can primitive circuits can be composed at scale, and into new capabilities? We answer both halves of this question: composition survives the jump from a hand-verified pair to dozens of independently-discovered circuits merged at once, and the resulting shared circuitry is able to do a more complex task that none of the 39 source circuits were individually pruned to solve. We do this by composing circuits pruned from an English phoneme classifier based on HuBERT, about as elementary a building block as a
spoken language offers, into an syllable classifier.

The outline of this paper is as follows. First, we identify failures of plain mask union (density domination, interference) and design a greedy, density-scaled algorithm that composes a pair of circuits.
Second, we generalize this to $N$ circuits with the help of a new data structure, \circuittree{}. Third,
we compose the resulting vowel and consonant \circuittree{}s into a
hierarchy-of-circuits pipeline with 99.5\% of matrices shared across phonemes. The resulting pipeline transcribes English CV and VC syllables, a task the individual circuits were not trained for, and which the composition algorithm does not optimize for either. We reach 64.8\% accuracy against the ceiling of 66.0\% obtained by running the circuits independently.

\section{Background on Circuit Discovery}
Circuit-discovery methods split along \emph{what} they prune. \textbf{Weight-level}
methods learn a differentiable mask directly over model parameters, e.g., via hard-concrete $L_0$ relaxations \cite{louizos2018learning} or movement
pruning \cite{sanh2020movement}, yielding a sparse sub-network but leaving
the dense computation graph untouched.
\textbf{Edge-level} methods instead prune the residual-stream computation
graph itself while keeping weights dense: ACDC searches over activation
patches \cite{conmy2023acdc}, and Edge Pruning \cite{bhaskar2024edgepruning}
learns a differentiable mask over edges with a straight-through relaxation. DiscoGP\cite{yu2025sheaf} is a circuit discovery algorithm that does both weight and edge pruning.

\section{Experimental Setup}

\textbf{Dataset.} We train and evaluate on the
\textbf{Articulatory Index} corpus \cite{schatz2015articulation},
originally collected by \cite{fousek2004new}, which contains pronunciations
of 10{,}329 CV and VC monosyllables produced by 20 English speakers (12
male, 8 female). The data for each phoneme is split into train, validation, and
test sets.  Results reported are for test. Thresholds for ensemble classifiers described in Section \ref{sec:pairwise} are tuned on the validation set.
All accept/reject decisions in Algorithm \ref{alg:pairwise} and its $N$-way
extension use tolerance $\varepsilon=0.02$ absolute F1; scale factors
(Eq.~\ref{eq:scale-n}) are clipped to $[0.05,5.0]$; the syllable type gate
(\S\ref{sec:hierarchy}) is itself a \discogp{} circuit under the same protocol.

Following \cite{wei2026hubertcircuits}, we prune 15 vowel and 24 consonant
\discogp{} circuits from HuBERT-base \cite{hsu2021hubert} with a frozen mean-pooled linear head
trained on the binary task of identifying whether a given vowel or consonant in present the syllable.
Below, we combine these binary circuits into ensemble multi-class classifiers as described in
Section \ref{sec:pairwise}.

The resulting weight circuits consist of 144 logit tensors (12 layers $\times$ \{Q,K,V,O,MLP-in,
MLP-out\} $\times$ \{weight,bias\}). Edge circuits are 49 logit tensors,
one per downstream node, each holding one logit per eligible upstream head
or MLP.

\section{Pairwise Circuit Composition}
\label{sec:pairwise}

We start by developing algorithms for composing two phonemes'
circuits $p_i,p_j$. We combine pairs of binary classifiers into three-way classifiers ${p_i,p_j,\mathrm{other}}$. Each binary classifier for phoneme $p_i$ is associated with a threshold $\tau_i$, tuned by validation set. When combining two classifiers, whichever phoneme's confidence surpasses its threshold the most will be the prediction. If neither does, we predict ``other''. In this setting, the classifiers being combined can be run independently, or can be composed into a single system with shared weights and edges.
 
\begin{figure*}[t!]
    \centering

    \begin{subfigure}[b]{0.32\textwidth}
        \centering
        \includegraphics[width=\linewidth]{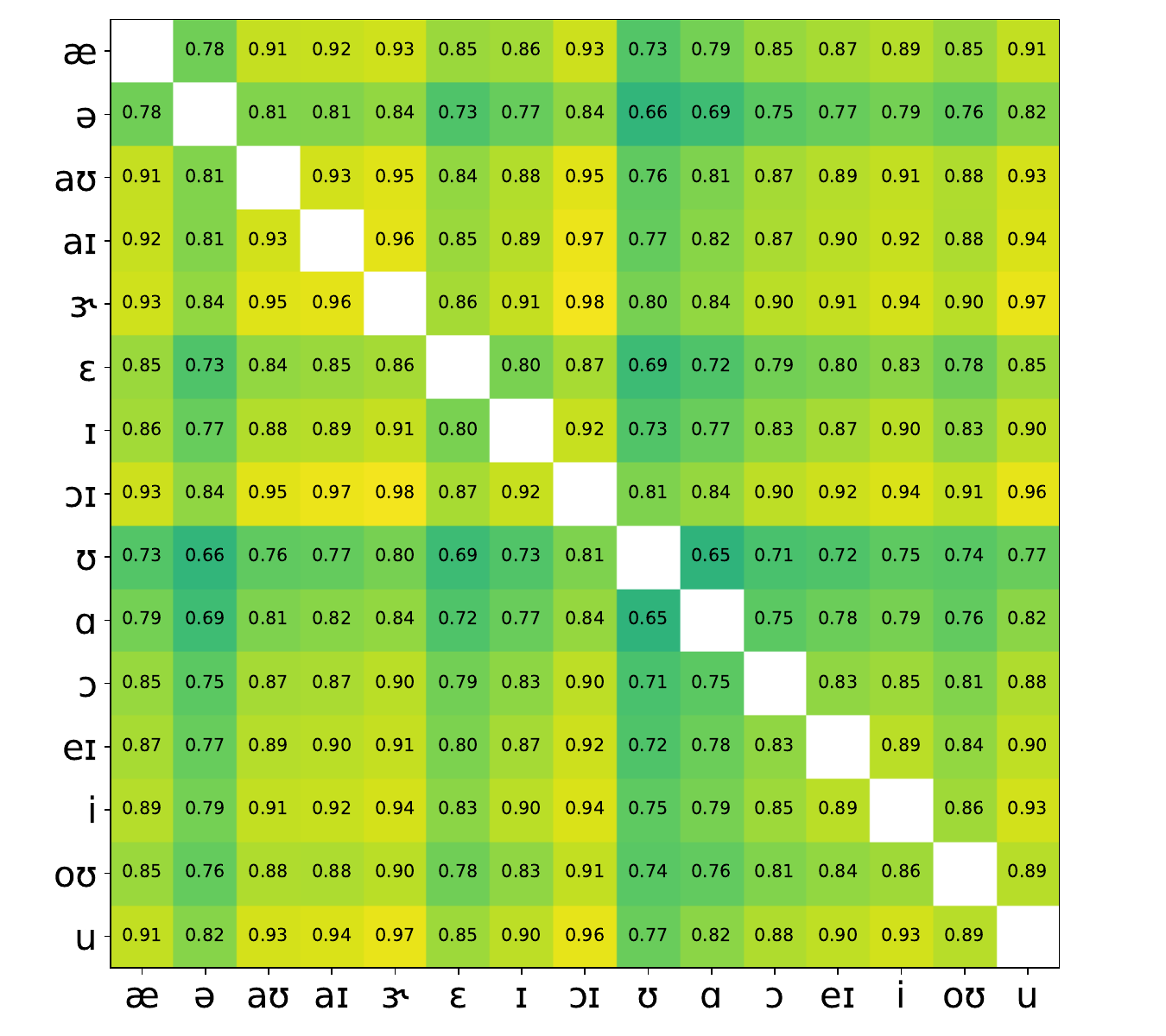}
        \caption{Independent circuits}
        \label{fig:heatmap_naive}
    \end{subfigure}\hfill
    \begin{subfigure}[b]{0.32\textwidth}
        \centering
        \includegraphics[width=\linewidth]{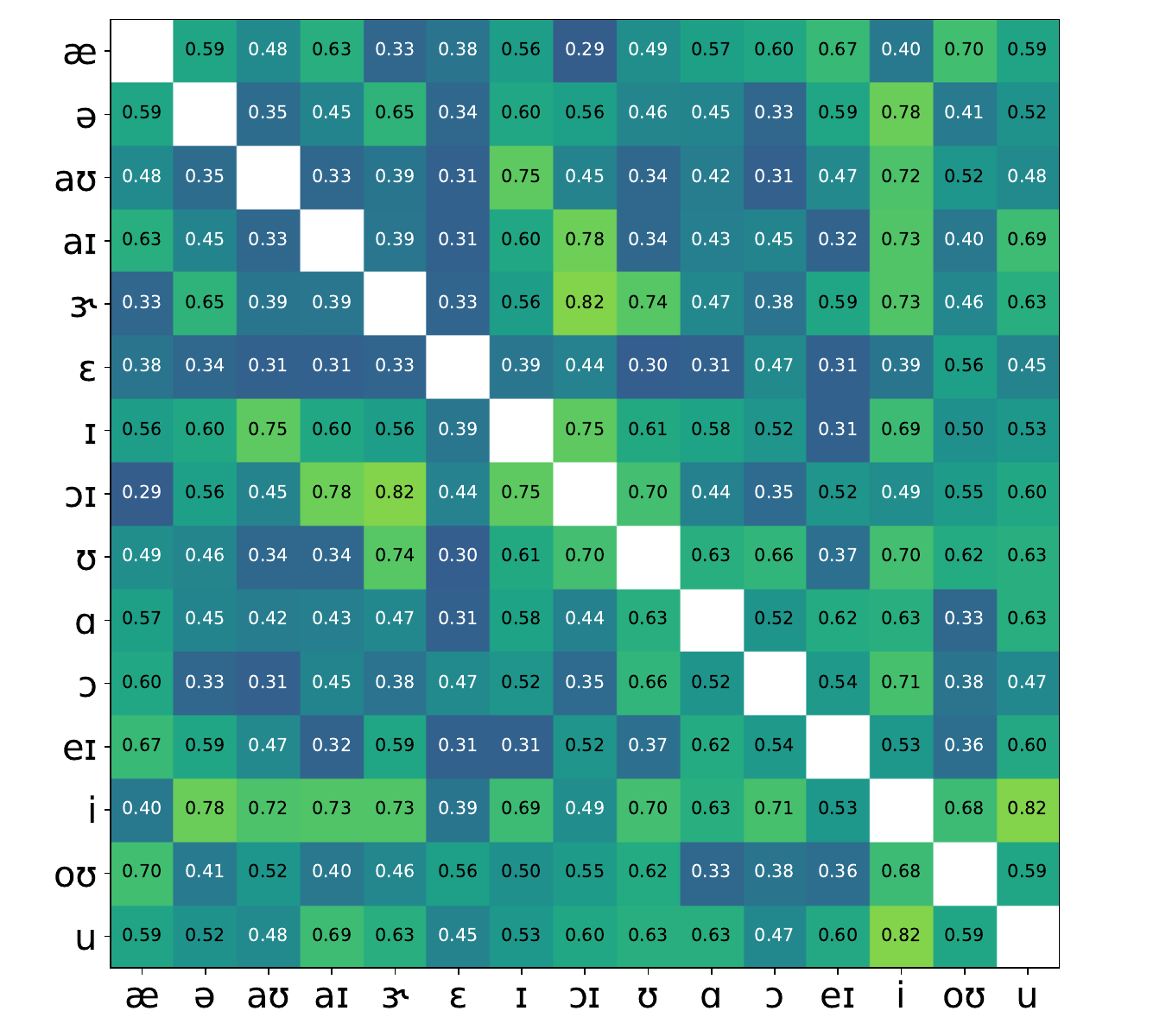}
        \caption{Naive union}
        \label{fig:heatmap_unmerged}
    \end{subfigure}\hfill
    \begin{subfigure}[b]{0.32\textwidth}
        \centering
        \includegraphics[width=\linewidth]{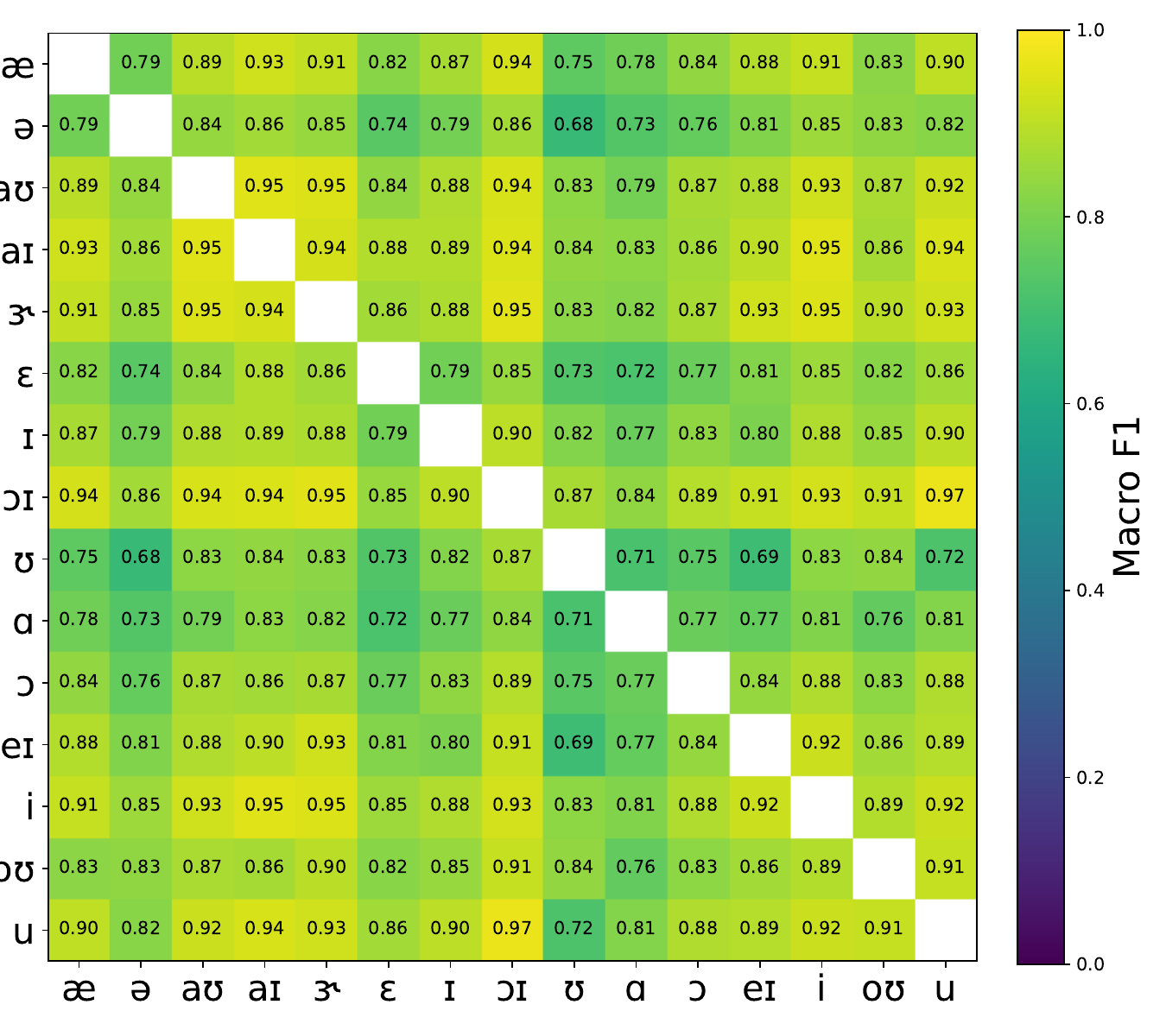}
        \caption{Circuit Composition (ours)}
        \label{fig:heatmap_merged}
    \end{subfigure}

    \caption{Pairwise macro-F1 heatmaps across all 15 vowels under three composition strategies: (a) naive union, (b) unmerged union, and (c) merged circuits.}
    \label{fig:heatmaps_all}
\end{figure*}

We first test a naive composition strategy by taking the element-wise union of weights and edges present in either circuit. As shown in Figure \ref{fig:heatmaps_all}, the independent, unmerged classifiers (Figure \ref{fig:heatmaps_all}a) substantially outperform this naive union across all vowel phoneme pairs (Figure \ref{fig:heatmaps_all}b). We attribute this performance degradation to two key issues: \emph{density domination} and \emph{interference}.
 


By \emph{density domination} we refer to the issue that circuits can have very
different weight densities (2--3\% for some phonemes, 15--20\% for others). Simple union is dominated by the denser circuit, so the composition
behaves like the dense circuit. We resolve this by scaling each mask in both circuits, or, in the general case, in all $N$ circuits.
For a shared key of the underlying model $k$, let $d_1(k),\dots,d_N(k)$
denote the fraction of active entries in each phoneme $p_1,\dots,p_N$'s 
mask. We calculate each phoneme's scale factor at key k $\sigma_i(k)$ (which have the same shape as circuit masks at key k) as follows:
\begin{equation}
\sigma_i(k) = \mathrm{clip}\big(\frac{t(k)}{d_i(k)},\,[s_{\min},s_{\max}]\big)
\label{eq:scale-n}
\end{equation}
where
\begin{equation}
t(k) = \Big(\prod_{n=1}^{N} d_n(k)\Big)^{1/N},
\end{equation}
is the geometric mean of the densities.

In following \textsc{Model Forward}() method where N circuits are applied to the HuBERT model, we first element-wise union (naive union) the N circuits and select the kept parameter/edge connection, then scale them by element-wise mulpilication of the scale factor $\sigma_i(k)$.
 

Even at matched density, overlaying two independently optimized masks
introduces circuit \emph{interference:} the presence of a given weight or
edge may be crucial to the behaviour of one circuit but detrimental to the other.
As the objects we are combining are binary 
masks, the techniques developed by \cite{yadav2023ties} for the case of model combination by weight averaging do not apply. We instead propose a greedy algorithm to merge a subset of 90\% the matrices that do not decrease the F1 score for each vowel more than some threshold.

Algorithm~\ref{alg:pairwise}
sorts the computation graph topologically and visits every key exactly once, proposing to take the union of the two circuits' masks for the given matrix, which we apply twice to the weight mask, then to the edge mask, only if both circuits' binary F1 stays within $\varepsilon$ of its independent (unmerged) performance.
 \begin{algorithm}[t]
\caption{Merge Clusters}
\label{alg:pairwise}
\begin{algorithmic}[1]
\small
\Require circuits $\mathcal{C}=\{C_n\}_{n=1}^{N}$, each $C_n=\{M_n(k)\}_{k\in K}$
a set of $|\mathrm{keys}|$ matrices; thresholds $\{\tau_n\}_{n=1}^{N}$; tolerance $\varepsilon$
\State $K\gets\textsc{Topological Sort}(\mathrm{keys})$
\State compute $\sigma_n(k)\ \forall k\in K,\ \forall n\in[N]$ via Eq.~\ref{eq:scale-n}
\State $f_n^0\gets\textsc{Model Forward}(C_n,\varnothing,\tau_n)\quad\forall n\in[N]$
\State $S\gets\varnothing$; $\Sigma\gets\varnothing$
\For{each $k\in K$} 
   \State $S'\gets S\cup\{k\}$
   \State $\hat{\mathcal{C}}\gets \mathcal{C}$ with masks unioned according to $S'$
   \State $\Sigma\gets\sigma_i(k)$
   \State $f_n\gets\textsc{Model Forward}(\hat{C_n},\Sigma,\tau_n)\quad\forall n\in[N]$ 
   \If{$f_n\ge f_n^0-\varepsilon$ \textbf{for all} $n\in[N]$}
      \State $S\gets S'$ \Comment{accepted}
   \EndIf
\EndFor
\State \Return $S,\ \{\widehat C_n|_S\}_{n=1}^{N},\ \Sigma|_S$
\end{algorithmic}
\end{algorithm}

Figure \ref{fig:heatmaps_all}c shows the performance is comparable to independent (unmerged) performance (Figure \ref{fig:heatmaps_all}a).

\section{ Tree-Merge Schedules and \circuittree{} Data Structure}
\label{sec:circuittree}
\begin{table}[t]
\centering

\small
\begin{tabular}{lccc}
\toprule
Schedule & Accuracy(\%) & Density(\%)\\
\midrule
Unmerged              & 82.3 & 1.13\\
Linear ($\Theta(kn^2)$)              & 77.9 & 1.11\\
Balanced ($\Theta(kn\log n)$)  & 81.8 & 1.10\\
Kruskal merge                       & 82.2 & 1.10\\
HuBERT Frozen~\cite{wei2026hubertcircuits}                       & 86.4 & 100\\
\bottomrule
\end{tabular}
\caption{15-way vowel accuracy of the $N$-way
merge schedules vs. Full HuBERT model, with post-merge circuit density.}
\label{tab:ablation}
\end{table}
Composing all 15 vowels (or 24 consonants) into a single module
still raises two further questions that a single invocation of Algorithm~\ref{alg:pairwise} cannot answer on its own. Assuming that we combine circuits iteratively, by adding one phoneme at a time, in what \emph{order} should
circuits be combined? And, when each new phoneme is added, how do we carry forward information about sharing validated at previous iterations?

\begin{algorithm}[t]
\caption{Linear schedule}
\label{alg:schedules-linear}
\begin{algorithmic}[1]
\footnotesize
\Function{Linear}{$C_1,\dots,C_n$}
  \State $T\gets C_1$
  \For{$v=2$ \textbf{to} $n$}\ $T\gets\textsc{MergeClusters}(T,C_v)$\ \EndFor
  \State \Return $T$
\EndFunction
\end{algorithmic}
\end{algorithm}
The simplest strategy composes $N$ phonemes one at a
time (Algorithm~\ref{alg:schedules-linear}). The trouble is
that step $i$ merges a single new member into a cluster that already holds
$i{-}1$ members, so \textsc{MergeClusters} (Algorithm~\ref{alg:pairwise})
must re-validate all $i$ resulting members against every one of the $k$
keys, costing $k i$ \textsc{ModelForward} calls. These costs accumulate
over the whole chain:
\begin{equation}
C_{\text{lin}} = k\sum_{i=2}^{n} i = k\Big(\tfrac{n(n+1)}{2}-1\Big) = \Theta(kn^2).
\label{eq:linear-cost}
\end{equation}

\begin{algorithm}[t]
\caption{Balanced schedule}
\label{alg:schedules-balanced}
\begin{algorithmic}[1]
\footnotesize
\Function{Balanced}{$C_1,\dots,C_n$}
  \State $Q\gets[C_1,\dots,C_n]$
  \While{$|Q|>1$}
     \State $Q'\gets[\,]$
     \For{$i=0,2,4,\dots$ while $i{+}1<|Q|$}
        \State\ $Q'.\mathrm{append}\big(\textsc{MergeClusters}(Q[i],Q[i{+}1])\big)$
     \EndFor
     \If{$|Q|$ odd}\ $Q'.\mathrm{append}(Q[\mathrm{last}])$\ \EndIf \Comment{carried forward}
     \State $Q\gets Q'$
  \EndWhile
  \State \Return $Q[0]$
\EndFunction
\end{algorithmic}
\end{algorithm}

A more efficient strategy avoids ever growing one
cluster far larger than its merge partner. Instead, singletons are merged
into pairs, pairs into fours, and so on (Algorithm~\ref{alg:schedules-balanced}). At round
$r=0,\dots,\log_2 n{-}1$, every cluster has size $2^r$, and there are
$n/2^{r+1}$ merges in that round, each invoking
\textsc{MergeClusters} on $2\cdot 2^r$ members at a cost of
$k\cdot 2^{r+1}$ \textsc{ModelForward} calls. The result is that
\emph{every round costs the same}, $kn$, regardless of $r$:
\begin{equation}
k\cdot\underbrace{\tfrac{n}{2^{r+1}}}_{\text{merges in round }r}\cdot\ \
\underbrace{2^{r+1}}_{\text{members validated / merge}} \;=\; k\,n.
\label{eq:balanced-round}
\end{equation}
Summing this constant cost, $kn$, over the $\log_2 n$ rounds:
\begin{equation}
C_{\text{bal}} = k\, n\log_2 n = \Theta(kn\log n).
\label{eq:balanced-cost}
\end{equation}

Both linear and balanced share a limitation: both are
\emph{blind} schedules, choosing merge order purely from cluster size and
never from how similar two phonemes actually are. We include a third
schedule as an upper bound on achievable performance. Here, the distance
between two phonemes is defined by the merge ratio from
\S\ref{sec:pairwise}, so that pairs sharing little of their masks are placed
far apart. 
Kruskal's algorithm is then
applied repeatedly, always composing the two closest clusters that have not
yet been joined. This schedule's complexity ranges from $\Theta(kn\log n)$ in
the best case to $\Theta(kn^2)$ in the worst case.
As  Table \ref{tab:ablation} shows, the balanced schedule matches the performance
of Kruskal merging while remaining $\Theta(kn\log n)$, making it preferred in practice.
 
\subsection{The \circuittree{} primitive}

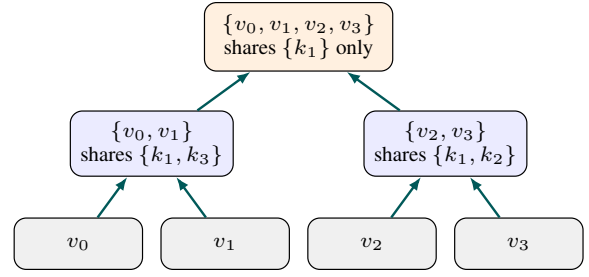
\begin{figure}[t!]
    \centering
    \resizebox{0.9\linewidth}{!}{
        \begin{tikzpicture}[scale=0.78,
          every node/.style={font=\scriptsize},
          leaf/.style={draw,rounded corners,minimum width=1.5cm,minimum height=0.6cm,align=center,fill=gray!12},
          node1/.style={draw,rounded corners,minimum width=1.9cm,minimum height=0.7cm,align=center,fill=blue!8},
          rootnode/.style={draw,rounded corners,minimum width=2.2cm,minimum height=0.8cm,align=center,fill=orange!12},
          edgeok/.style={-{Latex[length=1.6mm]},thick,color=teal!70!black},
        ]
          \node[leaf] (v0) at (-3.3,0) {$v_0$};
          \node[leaf] (v1) at (-1.1,0) {$v_1$};
          \node[leaf] (v2) at (1.1,0)  {$v_2$};
          \node[leaf] (v3) at (3.3,0)  {$v_3$};
          \node[node1] (n01) at (-2.2,1.5) {$\{v_0,v_1\}$\\shares $\{k_1,k_3\}$};
          \node[node1] (n23) at (2.2,1.5)  {$\{v_2,v_3\}$\\shares $\{k_1,k_2\}$};
          \node[rootnode] (root) at (0,3.1) {$\{v_0,v_1,v_2,v_3\}$\\shares $\{k_1\}$ only};
          \draw[edgeok] (v0)--(n01); \draw[edgeok] (v1)--(n01);
          \draw[edgeok] (v2)--(n23); \draw[edgeok] (v3)--(n23);
          \draw[edgeok] (n01)--(root); \draw[edgeok] (n23)--(root);
        \end{tikzpicture}
    }
    \caption{A \circuittree{} over four phonemes. $k_1$ validates at every level. $k_3$ only validated inside $\{v_0,v_1\}$: the root-level merge failed for $k_3$, but $v_0,v_1$ still use their shared value for it via Eq.~\ref{eq:resolve}, rather than reverting to private masks.}
    \label{fig:circuittree}
\end{figure}

\textsc{MergeClusters} operates on nodes of a \circuittree{}: a binary
tree whose leaves hold a phoneme's original circuit, and whose
internal nodes record \emph{only} which keys are shared among their two
children's members at that level (Figure \ref{fig:circuittree}).

To see why this structure is necessary, consider the natural alternative:
re-union the raw masks of all current members every round, and fall back to
fully private masks when any member's F1 drops too far. This
memoryless approach has a serious flaw. If a later round failed
by an unrelated third phoneme on some key $k$, the naive approach
discards \emph{all} sharing on $k$, including sharing between two
phonemes that had already validated it several rounds earlier.

\circuittree{} avoids this by never forgetting a validated result. Instead
of recomputing sharing from scratch, it resolves each member's effective
mask for key $k$ by walking down the tree and taking the first (i.e.\
closest-to-root) level at which that key was successfully shared, falling
back to the member's own leaf if it never was:
\begin{equation}
\footnotesize
\textsc{Resolve}(\mathcal T,m,k)=
\begin{cases}
U_{\mathcal T}(k) & k\in\mathrm{merged}(\mathcal T)\\
M_m(k) & \text{at a leaf}\\
\textsc{Resolve}(\mathrm{org}(\mathcal T,m),m,k) & \text{else}
\end{cases}
\normalsize
\label{eq:resolve}
\end{equation}
Scale factors are accumulated along the same path. Because sharing is
monotone non-decreasing along any root-to-leaf walk, a failed merge at the
root can never undo a key that two phonemes already agreed to share three
levels down.

\section{Hierarchy-of-Circuits Transcriber}
\label{sec:hierarchy}
We now compose the merged vowel and consonant \circuittree{} modules into a
syllable classifier. The pipeline has two stages. First, a separately-pruned
\discogp{} circuit acts as a \emph{gate}: it determines the syllable's
consonant/vowel order (e.g.\ CV vs.\ VC). Second, this order routes the
input to the corresponding vowel or consonant \circuittree{} module, which
resolves the specific phoneme. The two stages are combined by slotting the
predicted phonemes into the given template.

\begin{table}[t]
\centering
\caption{CV/VC-syllable transcription accuracy.}
\label{tab:hierarchy}
\small
\begin{tabular}{lccc}
\toprule
\multirow{2}{*}{System} & \multirow{2}{*}{Accuracy (\%)} & \multicolumn{2}{c}{Merge Ratio (\%)} \\
\cmidrule(lr){3-4}
 & & All & Partial \\
\midrule
Unmerged & 66.0 & 0.0 & 0.0 \\
\textbf{\circuittree{}} & \textbf{64.8} & 30.3 & 99.5 \\
\bottomrule
\end{tabular}
\end{table}

Table~\ref{tab:hierarchy} contains our central result. Collapsing 39
independent, binary phoneme circuits into just two shared \circuittree{}
modules costs only 1.2 points of accuracy (66.0\% $\to$ 64.8\%), while
99.5\% of keys are partially shared and 30.3\% are shared across all
phonemes. In other words, the vast majority of each phoneme's circuitry is
recovered from a common pool rather than stored independently, at a small
and controlled cost to accuracy.
\section{Conclusion}
This paper has demonstrated circuit composition at scale: a density-scaled,
interference-aware merge algorithm, generalized through \circuittree{} to
compose many circuits at once. Composing primitive
circuits for the vowels and consonants of English into a working
transcriber shows that primitive circuits can be joined well beyond a
handful, and into a capability none of them were individually pruned for. We believe this method is promising for much more complex problems, both in speech and in other domains.

\section{Compliance with Ethical Standards}

This study was conducted retrospectively using the Articulatory Index corpus, a pre-existing, publicly available speech dataset distributed by the Linguistic Data Consortium. No new data involving human or animal subjects were collected as part of this work. As this research involved secondary analysis of an existing, licensed open dataset rather than the collection of new human subject data, no additional ethical approval was required.
\bibliographystyle{IEEEbib}
\bibliography{strings,refs}

\end{document}